\documentclass{article}

\usepackage{arxiv}
\usepackage{amsmath}  
\usepackage{xcolor}   

\usepackage[utf8]{inputenc} 
\usepackage[T1]{fontenc}    
\usepackage{hyperref}       
\usepackage{url}            
\usepackage{booktabs}       
\usepackage{amsfonts}       
\usepackage{nicefrac}       
\usepackage{microtype}      
\usepackage{lipsum}
\usepackage{graphicx}
\graphicspath{ {./images/} }

\title{ConSens: Assessing context grounding in open-book question answering}

\author{
 Ivan Vankov \\
  Iris.ai BG, Sofia, Bulgaria\\
  INB, BAS, Sofia, Bulgaria\\
  \texttt{ivan@iris.ai} \\
   \And
 Matyo Ivanov \\
  Iris.ai BG, Sofia, Bulgaria\\
  \texttt{matyo@iris.ai} \\
  \And
 Adriana Correia \\
  Iris.ai, Oslo, Norway \\
  \texttt{adriana@iris.ai} \\
  \And
 Victor Botev \\
Iris.ai BG, Sofia, Bulgaria\\
  \texttt{victor@iris.ai} \\  
}

\begin{document}

\maketitle
\begin{abstract}
Large Language Models (LLMs) have demonstrated considerable success in open-book question answering (QA), where the task requires generating answers grounded in a provided external context. A critical challenge in open-book QA is to ensure that model responses are based on the provided context rather than its parametric knowledge, which can be outdated, incomplete, or incorrect. Existing evaluation methods, primarily based on the LLM-as-a-judge approach, face significant limitations, including biases, scalability issues, and dependence on costly external systems. 
To address these challenges, we propose a novel metric that contrasts the perplexity of the model response under two conditions: when the context is provided and when it is not. The resulting score quantifies the extent to which the model’s answer relies on the provided context. The validity of this metric is demonstrated through a series of experiments that show its effectiveness in identifying whether a given answer is grounded in the provided context. Unlike existing approaches, this metric is computationally efficient, interpretable, and adaptable to various use cases, offering a scalable and practical solution to assess context utilization in open-book QA systems.
\end{abstract}


\section{Introduction}
With the development of Large Language Models (LLMs), the promise of automation and revolution in the way we handle human knowledge has become closer to reality. An area still left for improvement is knowledge-intensive (KI) tasks. There are several requirements that have not yet been fulfilled to guarantee full trustworthiness in real-world scenarios. We will define here four such key requirements that, when fulfilled, will boost the adoption of LLMs for KI tasks in business cases dramatically:
\begin{itemize}
    \item \textbf{Utilizing the latest state of knowledge:} using up-to-date information that may not necessarily be present in the pre-trained data of the LLMs.
    \item \textbf{Handling of long-tail and rare knowledge:} many business cases require nuanced, domain specific or rare knowledge, which is crucial for accurate results (currently not present in pre-training data).
    \item \textbf{Handling diverse contexts:} many applications require domain-specific knowledge (e.g., medicine, finance, law) that is defined in common English but may have entirely different meanings in a specialized context. For instance, the phrase \textit{"Blue Roof"} might refer to "tiles reflecting heat" in a business scenario, but would simply be understood as "blue-colored tiles" by an LLM without the relevant context. 
    \item \textbf{Hallucination mitigation:} LLMs are prone to generate plausible but incorrect or unsupported by given information answers. Not only that but because of the high plausibility, they are also hard to flag as wrong and fix, making LLMs unreliable and hard to adopt for KI tasks. 
\end{itemize} 

Today, LLMs can address these requirements using two main approaches: weight modifying techniques \cite{robertsHowMuchKnowledge2020,zhangRAFTAdaptingLanguage2024} and input knowledge embedding techniques \cite{lewisRetrievalAugmentedGenerationKnowledgeIntensive2020,yangCuriousLLMElevatingMultiDocument2024}. Weight modification techniques are powerful, but lack scalability due to the need to keep up with new or frequently updated information. Modifying weights every time new information is ingested can be expensive and can lead to an uncontrolled degradation of the original abilities \cite{hawkinsEffectFinetuningLanguage2024,luoEmpiricalStudyCatastrophic2023}. For these reasons, these techniques become less practical for solving the business requirements defined above. Input knowledge embedding techniques like Retrieval Augmented Generation (RAG) embed external knowledge into the LLM input, which is significantly less expensive than weight modification, making it much more scalable and practical. Embedding knowledge in the input addresses all the requirements above, it gives the LLM the extra necessary information and context to answer the question and the desired focus to limit hallucinations. However, this method assumes that the LLMs are able to answer open-book questions - questions that are based on the presented knowledge and not on the pre-trained data. This makes open-book questions and answers (QA) evaluation an essential component for ensuring that LLMs can practically be useful for KI applications, and improve real-world implementations of LLMs for complex tasks.

The ability of an LLM to generate text that is grounded in the provided context is recognized as one of its defining characteristics \cite{jacoviFACTSGroundingLeaderboard2025,rashkinMeasuringAttributionNatural2021} and is particularly important for building open-book QA systems \cite{wuClashEvalQuantifyingTugofwar2024}. However, the factors and underlying mechanisms that determine whether a model will base its output on the provided context rather than on its parametric knowledge are complicated and still not entirely understood \cite{chenRichKnowledgeSources2022,phamWhosWhoLarge2024,xieAdaptiveChameleonStubborn2023,xuKnowledgeConflictsLLMs2024}. Given that there is no reliable way to predict whether the provided context will be used to guide the model output, it is important to have a way to determine this after the output is generated.

Our project is a competition on Kaggle (Predict Future Sales). We are provided with daily historical sales data (including each products’ sale date, block ,shop price and amount). And we will use it to forecast the total amount of each product sold next month. Because of the list of shops and products slightly changes every month. We need to create a robust model that can handle such situations.

\subsection{Contribution}
The contribution of the current work is threefold. First, we introduce ConSens, a novel metric for evaluating the influence of provided context on LLM-generated responses in open-book QA tasks. Second, we validate the effectiveness of ConSens through extensive experiments across diverse datasets and settings. Third, we demonstrate that ConSens is both cost-efficient and well-suited for real-time evaluation scenarios.

\subsection{Related work}
The traditional way to evaluate the quality of LLMs output is to compare it to a reference text (‘ground truth’) using metrics which quantify the degree of lexical, e.g. ROUGE \cite{lin_rouge_2004}, or semantic \cite{botev_word_2017,zhang_bertscore_2020} overlap. The same approach can be applied to identifying the role of the context on the generated answer in open-book QA by simply computing the similarity of the two texts (answer and context). Indeed, if the context was utilized by the LLM during generation then the resulting text should share lexical (e.g. named entities) and conceptual features with the context. The problem with metrics based on text similarity is that they tend to work at the surface level and struggle to account for the expressive abilities of modern LLMs \cite{gao_llm-based_2024}. More generally, similarity based metrics do not address the specifics of the open-book QA task such as how the relation between the answer and the context is conditioned on the question asked. 

The state-of-the-art method for determining the faithfulness (i.e. the level of 'grounding') of a model answer with respect to a provided context is to delegate the task to an external LLM \cite{esRAGASAutomatedEvaluation2023,jacoviFACTSGroundingLeaderboard2025,saad-falconARESAutomatedEvaluation2023}. For example, \cite{esRAGASAutomatedEvaluation2023} proposed a faithfulness metric that uses GPT-3.5 \cite{openaiGPT42024} to check whether all claims in the answer can be inferred from the context provided. A similar metric is also implemented in the Tonic Validate software development kit \cite{TonicValidate2023}. A more general approach is to define faithfulness as a custom evaluation criterion in general purpose LLM-as-a-judge evaluation frameworks such as GPTScore \cite{fuGPTScoreEvaluateYou2023} and Prometheus \cite{prometeusPaper}. The common characteristic of these metric implementations is that they rely on prompting an external, typically commercial, LLM accessed through an API. Consequently, they tend to be relatively slow and potentially costly, while also being influenced by LLM biases \cite{yeJusticePrejudiceQuantifying2024} and prompt sensitivity \cite{sclarQuantifyingLanguageModels2023}.

An alternative approach to quantifying faithfulness is to evaluate the conditional probability of generated tokens given the preceding context \cite{yuanBARTScoreEvaluatingGenerated2021,contextCite2024}. For example, \cite{contextCite2024} demonstrated that such a measure can be used to train a lightweight surrogate model for context attribution. We build on this idea by addressing some of the inherent problems of using token probabilities as a measure of model output quality and providing further evidence that manipulating the provided context can reveal its role in generating a given output.

\section{ConSens}
The goal of the ConSens metric is to provide a way to determine to what extent an LLM generated text is grounded in the context provided given a user query and a prompt. To do this, we contrast the perplexity of the model output in two conditions: when the context is provided ($P_C$) and when the context is empty ($P_E$). The ratio of these two perplexities, $P_E/P_C$, represents the effect of providing context on the probability of output tokens. The larger the ratio, the stronger the effect. 

The final value of ConSens is scaled to the interval [-1, 1] applying a sigmoid transformation:
\[
ConSens = \frac{2}{1 + e^{-r}} - 1, 
\]
 where
\[
r = \log \left( \frac{P_E}{P_C} \right).
\]
A value of ConSens close to 1 indicates that providing the context increases the likelihood of generating the given output. Values around and below 0 suggest that the context is not contributing to the output.

The perplexity of a generated text is computed by averaging the perplexity of the individual tokens:
\[
P_{\text{text}} = \frac{1}{N} \sum_{i=0}^N e^{-\log \left( p\left( \text{token}_i | \text{token}_{j<i}\right) \right)}
\]
, where $p(token_i|token_{j<i})$ is the probability of generating $token_i$ at step $i$ given a preceding sequence of tokens $token_{j<i}$.

\vskip 0.25in

\begin{table}[ht!]

\caption{Example of how the ConSens is calculated. The user question and the LLM generated answer are shown at the top. The words that are included in the calculation of the perplexity are underlined. “David Baker” and “is” are excluded because they also appear in the question and will result in relatively low perplexity regardless of the context condition. Closed set words such as ‘a’ (determiner), ‘and’ (conjunction) and punctuation marks are also excluded. The perplexities of the resulting three words are computed in four different contexts, starting with an empty one. The ConSens scores are given for the non-empty contexts only. 
}
\label{tab:perplexities}
\centering
\noindent\textit{\textbf{Question:}} What is David Baker known for? \\
\noindent\textit{\textbf{Answer:}} David Baker is a \underline{biochemist} and \underline{computational} \underline{biologist}. \\[1em]

\begin{tabular}{|p{3cm}|c|c|c|c|c|}
\hline
\multicolumn{1}{|c|}{Context} & \multicolumn{3}{c|}{Word perplexities} & \textbf{ConSens} \\ \hline

 & {\itshape "biochemist"} & {\itshape "computational"} & {\itshape "biologist"} &\\ \hline
(empty) & 4814.38 & 7117.1 & 1.61 & \\ \hline
David Baker is an American scientist who has pioneered methods to design proteins and predict their three-dimensional structures. & 263.73 & 293.92 & 1.72 & 0.91 \\ \hline
Colorless green ideas sleep furiously. & 3098.83 & 14517.0 & 2.01& -0.10 \\ \hline
David Baker is an English professional footballer. & 234191.27 & 61734.0 & 1.63 & -0.94 \\ \hline
\end{tabular}
\end{table}

\vskip 0.25in
In order to highlight the contrast between two conditions, we discard closed set words such as pronouns, determiners, and conjunctions (e.g., 'it', 'the', 'and'), as well as words which also appear in the user query. The rationale of filtering the set of words is to exclude words that are a priori likely to have similar perplexities in the two conditions and can thus lead to underestimating the effect of the provided context. Importantly, ConSens is not dependent on the absolute perplexity of the answer text, but rather on the ratio of its conditional perplexity in the two conditions. Therefore, ConSens is not directly affected by the problems of using perplexity as a general proxy of the qualities of generated text \cite{fangWhatWrongPerplexity2024,wangPerplexityPLMUnreliable2022}. A sample calculation of the metric is shown in Table \ref{tab:perplexities}.

\section{Validation}
We hereby present a series of experiments that aim to evaluate the performance of the proposed metric. We assume that the metric is used in a conventional open-book QA setting. In this setting, an LLM is tasked with responding to a user query given a context text (e.g., a list of retrieved documents). Each evaluation example consists of a query, a context, and a model output (answer). ConSens values are calculated using Llama 3.2 1B\footnote{https://huggingface.co/meta-llama/Llama-3.2-1B-Instruct} from the Meta LLama 3 family \cite{grattafioriLlama3Herd2024}. The choice of the evaluator model is justified by evidence that small models are more likely to adhere to the provided context than to respond using parametric knowledge \cite{biContextDPOAligningLanguage2024}. The following template was used for prompting the model during evaluation:


\begin{verbatim}
Consider the following context:
Context:
{{context}}
Please answer the following question:
{{question}}
Answer:
\end{verbatim}

\subsection{Experiment 1: Evaluation of grounded versus ungrounded answers.}
In the first experiment, we evaluated the ability of the metric to distinguish between answers that are based on the given context and those that are not by using the the publically available WikiEval dataset \cite{esRAGASAutomatedEvaluation2023}. Each example contains a question, a context (excerpt from Wikipedia), and two answers generated using an unspecified version of ChatGPT. The first answer (referred to as the ‘grounded’ answer) was generated by providing the corresponding context, while the second answer (‘ungrounded’) was based on parametric knowledge of the model; that is, the model did not have access to the context. For each example, we computed the value of the metric for each of the answers. 

\begin{figure}
\centering
\includegraphics[scale=0.8]{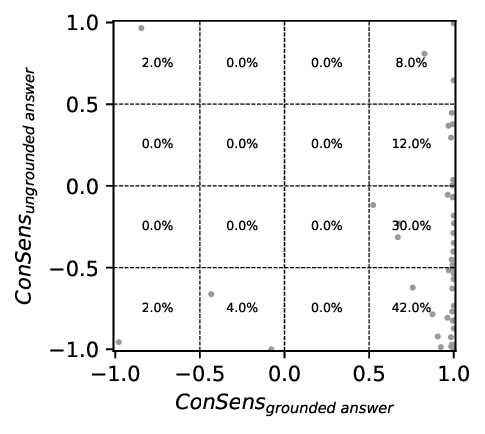}
\vspace*{-5mm}
\caption{Experiment 1 results. The x-axis represents the ConSens value for the grounded answer, while the y-axis corresponds to the ConSens value for the ungrounded answer. The percentage values indicate the proportion of scores that fall within the respective ranges of the two ConSens scores.}
\label{fig1}
\end{figure}

The results of Experiment 1 are presented in Fig.~\ref{fig1}. The distribution of the metric scores clearly shows that ConSens results in higher values when grounded answers are evaluated (m = 0.83, HDI\footnote{The highest density interval (HDI) stands for the shortest interval which contains a certain proportion of the data. The 90\% HDI is used in all cases in this article.}: [0.67, 1]) compared to ungrounded answers (m = -0.34, HDI: [-1.00, 0.45]). The average difference between conditions was 1.17 (HDI: [0.2, 2.00]). The value of the ROC AUC was 0.92 suggesting that the metric provides a reliable signal to differentiate answers that are based on the context provided from those that are not.

\subsection{Experiment 2: Evaluation of full versus partial context.}
The second experiment aims to assess the sensitivity of the metric to removing a critical section of the provided context. A new dataset was constructed using a publicly available collection of biomedical abstracts \cite{attalDatasetPlainLanguage2023}. For each of the abstracts, we prompted OpenAI gpt-4o to generate a question based on two different consecutive sentences from the abstract. The same model was then used to generate an answer to each of the questions using the entire abstract. The result of the procedure was a set of 657 questions paired with a full context (i.e. the whole abstract), partial context (the same abstract, but with the two critical sentences used to generate the question removed), and a model answer (generated by providing the full context).
The rationale for creating this dataset is to test the ability of the metric to distinguish the effect of two contexts on a generated answer when the contexts are very similar as far as topic and terminology go. The difference between the full and partial context condition is only in those two consecutive sentences that are used to generate the question (note that the answer is always generated using the full context). This setting makes a stronger test of the ConSens metric than the WikiEval data set used in Experiment 1. 

\begin{figure}
\centering
\includegraphics[scale=0.8]{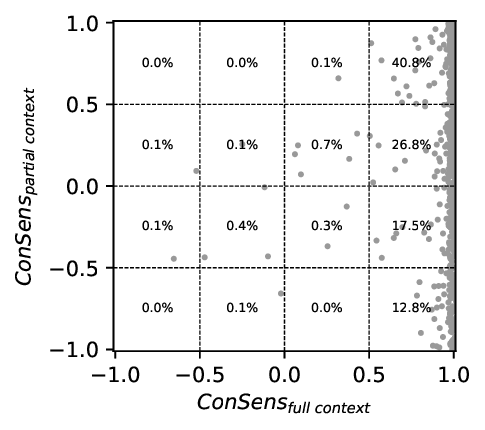}
\vspace*{-5mm}
\caption{Experiment 2 results. Values on the x axis represent the ConSens value when the full context was provided and the y axis represents the ConSens scores in the partial context condition.}
\label{fig2}
\end{figure}

The results (Fig.~\ref{fig2}) indicate that ConSens maxed out when the full context is provided. The value of ConSens was reliably higher in the full context condition (m = 0.95, HDI: [0.91, 1.0]) than in the partial context condition (m = 0.27, HDI: [-0.63, 1.0]). The mean difference between conditions was 0.69 (HDI: [0.0, 1.75]) and the ROC AUC was 0.93. Compared to the results of Experiment 1, there was an increased number of cases in which the metric resulted in similar scores, reflecting the increased difficulty to differentiate the effects of the full context and the partial context on the answer.

\subsection{Experiment 3: Context grounding in RAG setting.}
The goal of the last experiment was to validate the use of ConSens in a typical RAG setting, in which the context consists of more than one document. The dataset consisted of the the same questions and documents as in Experiment 2. The abstracts were vectorized using sentence embeddings\footnote{https://huggingface.co/sentence-transformers/all-MiniLM-L6-v2} and fed into a vector database\footnote{https://github.com/chroma-core/chroma}. For each example, the question was used to retrieve the three most similar documents from the database. The context was constructed by concatenating the retrieved documents in randomized order. The answer was generated by providing the question and the context (consisting of the three documents) to gpt-4o.

\begin{figure}
\centering
\includegraphics[scale=0.8]{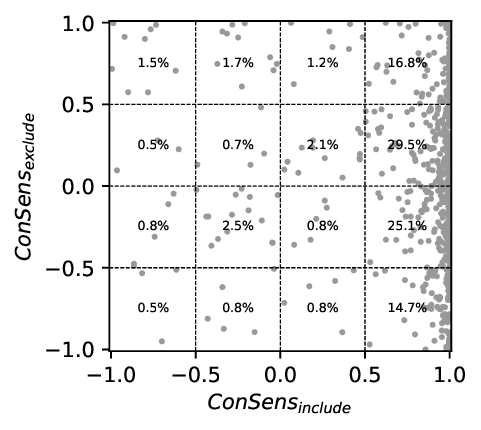}

\caption{Experiment 3 results. The x axis represents the minimum value of ConSens when the correct document (i.e. the abstract which was used to generate the question) was included in the context. Accordingly, the y axis stands for the value of ConSens when the correct document was not part of the context.}
\label{fig3}
\end{figure}

Only the retrieval hits were considered for evaluating ConSens, i.e. the cases in which the correct document (i.e., the abstract which was used to generate the question) was retrieved, which amounted to 88.21\% of the examples. The validity of ConSens was tested by checking whether it can be used to identify which of the three retrieved documents was the most influential on the response. To do this, we manipulated the context by repeatedly excluding each of the documents from it and calculated the metric for each of the resulting three contexts. The idea of this manipulation is to examine the sensitivity of the metric to the exclusion of the most relevant context segment, compared to the other two. 

The results of Experiment 3 are presented in Fig.~\ref{fig3}. Removing the correct context resulted in the lowest ConSens score in 89.43\% of the cases. Including the correct document in the context lead to a higher score (m = 0.77, HDI: [0.11, 1]) compared to when it was not included (m = 0.06, HDI: [-0.67, 1]). The mean difference between conditions was 0.72 (HDI: [-0.08, 1.98]). The ROC AUC was 0.88. 

\subsection{Comparison with other metrics}
In order to gain further insight into the usefulness of the ConSens metric, the three experiments were repeated with the LLM-as-a-judge metric `answer consistency' available in the Tonic Validate LLM/RAG evaluation framework \cite{TonicValidate2023}. The consistency score of the answers is calculated using OpenAI 'gpt-4-turbo-preview' to create a bulleted list of the main points in the response text and check the proportion of points that can be attributed to the provided context. The implementation of this metric therefore results in multiple calls to the evaluator model, the exact number depending on the number of main points extracted.

Furthermore, we also checked whether the performance of a ConSens across the three experiments can be explained in terms of the similarity between the context and the answer texts. To do this, we computed the cosine similarity of the vector embeddings of the two texts using the OpenAI 'text embedding-3-large' model. 

The resulting ROC AUC scores for each of the metrics (Table \ref{tab:tonic}) demonstrate that the performance of ConSens is comparable and, in some cases, superior to a much slower and more expensive LLM-as-a-judge metric and cannot be attributed to the mere similarity between the answer and the context texts.

\begin{table}
\centering
\vspace{10pt}
\setlength{\tabcolsep}{4pt}
\caption{ROC AUC scores across the three experiments.}
\label{tab:tonic}
\begin{tabular}{|l|c|c|c|}
\hline
\textbf{} & ConSens & Tonic Answer Consistency & Answer-Context Similarity\\
\hline
Experiment 1 & {\bfseries 0.92} & {\bfseries 0.92} & 0.62\\ \hline
Experiment 2 & {\bfseries 0.93} & 0.75 & 0.68\\ \hline
Experiment 3 & 0.88 & {\bfseries 0.91} & 0.76\\ \hline
\end{tabular}
\end{table}
\vspace{10pt}

\subsection{Evaluator model}

All the ConSens scores presented thus far have been based on token perplexity calculations using Llama 3.2 1B. We selected this model because it is one of the smallest industrial-grade models capable of following instructions and integrating information from relatively large texts. To demonstrate that ConSense's performance is not dependent on this specific model, we repeated the three experiments using three other models from the same family. The results, shown in Table \ref{tab:evaluators}, indicate that ConSense's performance remained largely independent from the choice of evaluator model. In other words, employing a larger (and slower and more expensive to deploy) model did not significantly impact ConSense’s ability to distinguish between grounded and ungrounded generations of LLM.

\begin{table}
\centering
\vspace{10pt}
\setlength{\tabcolsep}{4pt}
\caption{ConSens ROC AUC scores as a function of the evaluator model.} 
\label{tab:evaluators}
\begin{tabular}{|l|c|c|c|c|}
\hline
\textbf{} & Llama 3.2 1B & Llama 3.2 3B & Llama 3.1 8B & Llama 3.1 70B \\
\hline
Experiment 1 & 0.92 & {\bfseries 0.97} & 0.93 & 0.92 \\ \hline
Experiment 2 & 0.93 & {\bfseries 0.94} & 0.93 & 0.92 \\ \hline
Experiment 3 & {\bfseries 0.88} & 0.87 & 0.86 & 0.82 \\ \hline
\end{tabular}
\vspace{10pt}
\end{table}

\section{Discussion}

The pattern of results in all three experiments supports the validity of the ConSens metric. In Experiment 1, we show that ConSens can distinguish answers which are based on a given context from those which are not. In Experiment 2, the metric is used to determine which context is more likely to lead to generating a fixed answer. Finally, in Experiment 3, we demonstrate that ConSens can be used in RAG systems to identify which part of a given context contributes the most to a generated answer. It is worth noting that Experiments 2 and 3 are based on a dataset which has been constructed to be especially challenging for the metric. Importantly, the observed distribution of the ConSens scores between experiments and conditions indicates that the precision of the metric is high enough to be practically useful in real-world applications.

Although ConSens seems to be a very promising approach to the assessment of the performance of open-book QA systems, it also has certain limitations. For example, it requires access to the raw outputs (i.e. logits) of the model used for evaluation. This makes it difficult to set up with an evaluation LLM accessible via API only. However, in this paper we demonstrate that the metric works well even with one of the most lightweight LLMs available, meaning that it can be easily and efficiently deployed on-premise on a wide range of platforms, including edge and mobile devices. Calculating token perplexities can be also be parallelized, meaning that the metric scores can be obtained with a single pass through the model even when we need to assess several texts per example, as was the case in Experiment 3. 

Another limitation is that the metric scale is non-linear, which means that care must be taken when drawing conclusions from its data. For example, the metric cannot be used to show that the effect of document A on the answer is twice the effect of document B or that the difference between the effects of document A and B is the same as the difference between the effects of document C and D.

The idea of ConSens gives a lot of room for improvement and possible other use cases. For example, the same approach can be used to assess whether an auto-generated summary adequately covers all parts of the input. Another potential application is identifying critical segments of an LLM prompt that significantly influence the model's output. More broadly, the metric is applicable in any open-book QA scenario where the goal is to evaluate how the output of a generative LLM is conditioned on a specific segment of its input.

\section{Conclusions}

In this work, we introduce the ConSens metric as a tool to quantify how strongly the output of an LLM is influenced by the provided context. We demonstrate its effectiveness in practical business applications, such as retrieval-augmented generation (RAG), and highlight its versatility for other tasks. In addition, we proved several important characteristics that set it apart from existing approaches. 

Unlike other evaluation methods based on LLM-as-a-judge approach, ConSens does not require access to state-of-the-art LLM APIs, making it easy to deploy internally while ensuring security and privacy. It is computationally efficient, as it can be applied using any lightweight models and not necessarily the model generating the responses under evaluation. Furthermore, it is not dependent on model-specific prompt engineering, allowing it to work seamlessly with any sufficiently capable LLM. Finally, it is based on a simple and intuitive idea, making it easier to interpret and implement. Therefore we believe that ConSens makes an important step in the development of tools and assets for evaluating open-book question and answer systems in knowledge-intensive tasks.

\bibliographystyle{splncs04}
\bibliography{paper}

\begin{thebibliography}{10}
\providecommand{\url}[1]{\texttt{#1}}
\providecommand{\urlprefix}{URL }
\providecommand{\doi}[1]{https://doi.org/#1}

\bibitem{attalDatasetPlainLanguage2023}
Attal, K., Ondov, B., Demner-Fushman, D.: A dataset for plain language adaptation of biomedical abstracts. Sci Data  \textbf{10}(1), ~8 (2023), \url{https://www.nature.com/articles/s41597-022-01920-3}

\bibitem{biContextDPOAligningLanguage2024}
Bi, B., Huang, S., Wang, Y., Yang, T., Zhang, Z., Huang, H., Mei, L., Fang, J., Li, Z., Wei, F., Deng, W., Sun, F., Zhang, Q., Liu, S.: Context-dpo: Aligning language models for context-faithfulness. CoRR  (2024), \url{https://doi.org/10.48550/arXiv.2412.15280}

\bibitem{botev_word_2017}
Botev, V., Marinov, K., Schäfer, F.: Word importance-based similarity of documents metric ({WISDM}): Fast and scalable document similarity metric for analysis of scientific documents. In: Proceedings of the 6th International Workshop on Mining Scientific Publications. pp. 17--23. {ACM} (2017), \url{https://dl.acm.org/doi/10.1145/3127526.3127530}

\bibitem{chenRichKnowledgeSources2022}
Chen, H.T., Zhang, M.J.Q., Choi, E.: Rich knowledge sources bring complex knowledge conflicts: Recalibrating models to reflect conflicting evidence. In: Proceedings of the 2022 EMNLP, {EMNLP} 2022, Abu Dhabi, United Arab Emirates, December 7-11, 2022. pp. 2292--2307. ACL (2022), \url{https://doi.org/10.18653/v1/2022.emnlp-main.146}

\bibitem{contextCite2024}
Cohen-Wang, B., Shah, H. abd~Georgiev, K., Madry, A.: Contextcite: Attributing model generation to context (2024), \url{http://arxiv.org/abs/1904.09675}

\bibitem{esRAGASAutomatedEvaluation2023}
ES, S., James, J., Anke, L.E., Schockaert, S.: Ragas: Automated evaluation of retrieval augmented generation. In: Proceedings of the 18th EACL, {EACL} 2024 - System Demonstrations, St. Julians, Malta, March 17-22, 2024. pp. 150--158. ACL (2024), \url{https://aclanthology.org/2024.eacl-demo.16}

\bibitem{fangWhatWrongPerplexity2024}
Fang, L., Wang, Y., Liu, Z., Zhang, C., Jegelka, S., Gao, J., Ding, B., Wang, Y.: What is wrong with perplexity for long-context language modeling? CoRR  (2024), \url{https://doi.org/10.48550/arXiv.2410.23771}

\bibitem{fuGPTScoreEvaluateYou2023}
Fu, J., Ng, S., Jiang, Z., Liu, P.: Gptscore: Evaluate as you desire. In: Proceedings of the 2024 NAACL (Volume 1: Long Papers), {NAACL} 2024, Mexico City, Mexico, June 16-21, 2024. pp. 6556--6576. ACL (2024), \url{https://doi.org/10.18653/v1/2024.naacl-long.365}

\bibitem{gao_llm-based_2024}
Gao, M., Hu, X., Ruan, J., Pu, X., Wan, X.: {LLM}-based {NLG} evaluation: Current status and challenges (2024), \url{http://arxiv.org/abs/2402.01383}

\bibitem{hawkinsEffectFinetuningLanguage2024}
Hawkins, W., Mittelstadt, B.D., Russell, C.: The effect of fine-tuning on language model toxicity. CoRR  (2024), \url{https://doi.org/10.48550/arXiv.2410.15821}

\bibitem{jacoviFACTSGroundingLeaderboard2025}
Jacovi, A., Wang, A., Alberti, C., Tao, C., Lipovetz, J., Olszewska, K., Haas, L., Liu, M., Keating, N., Bloniarz, A., Saroufim, C., Fry, C., Marcus, D., Kukliansky, D., Tomar, G.S., Swirhun, J., Xing, J., Wang, L., Gurumurthy, M., Aaron, M., Ambar, M., Fellinger, R., Wang, R., Zhang, Z., Goldshtein, S., Das, D.: The {{FACTS Grounding Leaderboard}}: {{Benchmarking LLMs}}' {{Ability}} to {{Ground Responses}} to {{Long-Form Input}} (2025), \url{https://arxiv.org/abs/2501.03200}

\bibitem{prometeusPaper}
Kim, S., Suk, J., Longpre, S., Lin, B.Y., Shin, J., Welleck, S., Neubig, G., Lee, M., Lee, K., Seo, M.: Prometheus 2: An open source language model specialized in evaluating other language models. In: Proceedings of the 2024 EMNLP (EMNLP). pp. 4334--4353. Miami, Florida, USA (2024). \doi{10.18653/v1/2024.emnlp-main.248}

\bibitem{lewisRetrievalAugmentedGenerationKnowledgeIntensive2020}
Lewis, P.S.H., Perez, E., Piktus, A., Petroni, F., Karpukhin, V., Goyal, N., K{\"{u}}ttler, H., Lewis, M., Yih, W., Rockt{\"{a}}schel, T., Riedel, S., Kiela, D.: Retrieval-augmented generation for knowledge-intensive {NLP} tasks. In: Advances in Neural Information Processing Systems 33: NeurIPS 2020 (2020), \url{https://proceedings.neurips.cc/paper/2020/hash/6b493230205f780e1bc26945df7481e5-Abstract.html}

\bibitem{lin_rouge_2004}
Lin, C.Y.: {ROUGE}: A package for automatic evaluation of summaries. In: Text Summarization Branches Out. pp. 74--81. ACL (2004), \url{https://aclanthology.org/W04-1013/}

\bibitem{luoEmpiricalStudyCatastrophic2023}
Luo, Y., Yang, Z., Meng, F., Li, Y., Zhou, J., Zhang, Y.: An empirical study of catastrophic forgetting in large language models during continual fine-tuning. CoRR  (2023), \url{https://doi.org/10.48550/arXiv.2308.08747}

\bibitem{grattafioriLlama3Herd2024}
Meta, L.T.A..: The llama 3 herd of models. CoRR  (2024), \url{https://doi.org/10.48550/arXiv.2407.21783}

\bibitem{openaiGPT42024}
OpenAI: Gpt-4 technical report (2024), \url{https://arxiv.org/abs/2303.08774}

\bibitem{phamWhosWhoLarge2024}
Pham, Q., Ngo, H., Luu, A.T., Nguyen, D.Q.: Who's who: Large language models meet knowledge conflicts in practice. In: Findings of the Association for Computational Linguistics: {EMNLP} 2024, Miami, Florida, USA, November 12-16, 2024. pp. 10142--10151. ACL (2024), \url{https://aclanthology.org/2024.findings-emnlp.593}

\bibitem{rashkinMeasuringAttributionNatural2021}
Rashkin, H., Nikolaev, V., Lamm, M., Aroyo, L., Collins, M., Das, D., Petrov, S., Tomar, G.S., Turc, I., Reitter, D.: Measuring attribution in natural language generation models. Comput. Linguistics  \textbf{49}(4),  777--840 (2023), \url{https://doi.org/10.1162/coli\_a\_00486}

\bibitem{robertsHowMuchKnowledge2020}
Roberts, A., Raffel, C., Shazeer, N.: How much knowledge can you pack into the parameters of a language model? In: Proceedings of the 2020 EMNLP, {EMNLP} 2020, Online, November 16-20, 2020. pp. 5418--5426. ACL (2020), \url{https://doi.org/10.18653/v1/2020.emnlp-main.437}

\bibitem{saad-falconARESAutomatedEvaluation2023}
Saad{-}Falcon, J., Khattab, O., Potts, C., Zaharia, M.: {ARES:} an automated evaluation framework for retrieval-augmented generation systems. In: Proceedings of the 2024 NAACL (Volume 1: Long Papers), {NAACL} 2024, Mexico City, Mexico, June 16-21, 2024. pp. 338--354. ACL (2024), \url{https://doi.org/10.18653/v1/2024.naacl-long.20}

\bibitem{sclarQuantifyingLanguageModels2023}
Sclar, M., Choi, Y., Tsvetkov, Y., Suhr, A.: Quantifying language models' sensitivity to spurious features in prompt design or: How {I} learned to start worrying about prompt formatting. In: The Twelfth ICLR, {ICLR} 2024, Vienna, Austria, May 7-11, 2024. OpenReview.net (2024), \url{https://openreview.net/forum?id=RIu5lyNXjT}

\bibitem{TonicValidate2023}
Tonic.ai: {T}onic {V}alidate guide | {T}onic {V}alidate --- docs.tonic.ai. \url{https://docs.tonic.ai/validate} (2023), [Accessed 28-03-2025]

\bibitem{wangPerplexityPLMUnreliable2022}
Wang, Y., Deng, J., Sun, A., Meng, X.: Perplexity from {PLM} is unreliable for evaluating text quality. CoRR  (2022), \url{https://doi.org/10.48550/arXiv.2210.05892}

\bibitem{wuClashEvalQuantifyingTugofwar2024}
Wu, K., Wu, E., Zou, J.: {{ClashEval}}: {{Quantifying}} the tug-of-war between an {{LLM}}'s internal prior and external evidence (2024), \url{https://arxiv.org/abs/2404.10198}

\bibitem{xieAdaptiveChameleonStubborn2023}
Xie, J., Zhang, K., Chen, J., Lou, R., Su, Y.: Adaptive chameleon or stubborn sloth: Revealing the behavior of large language models in knowledge conflicts. In: The Twelfth International Conference on Learning Representations, {ICLR} 2024, Vienna, Austria, May 7-11, 2024. OpenReview.net (2024), \url{https://openreview.net/forum?id=auKAUJZMO6}

\bibitem{xuKnowledgeConflictsLLMs2024}
Xu, R., Qi, Z., Guo, Z., Wang, C., Wang, H., Zhang, Y., Xu, W.: Knowledge conflicts for llms: {A} survey. In: Proceedings of the 2024 EMNLP, {EMNLP} 2024, Miami, FL, USA, November 12-16, 2024. pp. 8541--8565. ACL (2024), \url{https://aclanthology.org/2024.emnlp-main.486}

\bibitem{yangCuriousLLMElevatingMultiDocument2024}
Yang, Z., Zhu, Z.: Curiousllm: Elevating multi-document {QA} with reasoning-infused knowledge graph prompting. CoRR  (2024), \url{https://doi.org/10.48550/arXiv.2404.09077}

\bibitem{yeJusticePrejudiceQuantifying2024}
Ye, J., Wang, Y., Huang, Y., Chen, D., Zhang, Q., Moniz, N., Gao, T., Geyer, W., Huang, C., Chen, P., Chawla, N.V., Zhang, X.: Justice or prejudice? quantifying biases in llm-as-a-judge. CoRR  (2024), \url{https://doi.org/10.48550/arXiv.2410.02736}

\bibitem{yuanBARTScoreEvaluatingGenerated2021}
Yuan, W., Neubig, G., Liu, P.: Bartscore: Evaluating generated text as text generation. In: Advances in Neural Information Processing Systems 34: NeurIPS 2021, NeurIPS 2021, December 6-14, 2021, virtual. pp. 27263--27277 (2021), \url{https://proceedings.neurips.cc/paper/2021/hash/e4d2b6e6fdeca3e60e0f1a62fee3d9dd-Abstract.html}

\bibitem{zhangRAFTAdaptingLanguage2024}
Zhang, T., Patil, S.G., Jain, N., Shen, S., Zaharia, M., Stoica, I., Gonzalez, J.E.: {RAFT:} adapting language model to domain specific {RAG}. CoRR  (2024), \url{https://doi.org/10.48550/arXiv.2403.10131}

\bibitem{zhang_bertscore_2020}
Zhang, T., Kishore, V., Wu, F., Weinberger, K.Q., Artzi, Y.: {BERTScore}: Evaluating text generation with {BERT} (2020), \url{http://arxiv.org/abs/1904.09675}

\end{thebibliography}

\end{document}